\documentclass[runningheads]{llncs}
\usepackage{graphicx}

\usepackage{tikz}
\usepackage{comment}
\usepackage{amsmath,amssymb} 
\usepackage{color}
\usepackage{caption}
\usepackage{multirow}
\usepackage{makecell}
\usepackage{amsmath}
\usepackage{amssymb}
\usepackage{subfigure}
\usepackage{footmisc}
\usepackage{epstopdf}

\begin{document}
\pagestyle{headings}
\mainmatter
\def\ECCVSubNumber{4174}  

\title{NAS-Count: Counting-by-Density with Neural Architecture Search} 

\titlerunning{NAS-Count: Counting-by-Density with Neural Architecture Search}
%
\author{Yutao Hu \inst{1} \protect\footnotemark[1] \and
Xiaolong Jiang \inst{4} \protect\footnotemark[1] \and
Xuhui Liu\inst{1} \and
Baochang Zhang\inst{5} \and
\\Jungong Han\inst{6} \and
Xianbin Cao\inst{1,2,3} \protect\footnotemark[2] \and
David Doermann\inst{7}}
\authorrunning{Y. Hu {\it et al.}}
%
\institute{School of Electronic and Information Engineering,\\ Beihang University, Beijing, China \and
Key Laboratory of Advanced Technologies for Near Space Information Systems, Ministry of Industry and Information Technology of China \and
Beijing Advanced Innovation Center for Big Data-Based Precision Medicine,China \and
YouKu Cognitive and Intelligent Lab, Alibaba Group \and
Beihang University, Beijing, China \and
Computer Science Department, Aberystwyth University, SY23 3FL, UK \and
Department of Computer Science and Engineering,\\ University at Buffalo, New York, USA \\
\email{\{huyutao, bczhang, xbcao\}@buaa.edu.cn, xainglu.jxl@alibaba-inc.com, xuhui\_cc@126.com, jungonghan77@gmail.com, doermann@buffalo.edu}
}
\maketitle
\renewcommand{\thefootnote}{\fnsymbol{footnote}}
\footnotetext[1]{Contribute equally}
\footnotetext[2]{Corresponding author}
\begin{abstract}
Most of the recent advances in crowd counting have evolved from hand-designed density estimation networks, where multi-scale features are leveraged to address the scale variation problem, but at the expense of demanding design efforts. In this work, we automate the design of counting models with Neural Architecture Search (NAS) and introduce an end-to-end searched encoder-decoder architecture, Automatic Multi-Scale Network (AMSNet). Specifically, we utilize a counting-specific two-level search space. The encoder and decoder in AMSNet are composed of different cells discovered from micro-level search, while the multi-path architecture is explored through macro-level search. To solve the pixel-level isolation issue in MSE loss, AMSNet is optimized with an auto-searched Scale Pyramid Pooling Loss (SPPLoss) that supervises the multi-scale structural information. Extensive experiments on four datasets show AMSNet produces state-of-the-art results that outperform hand-designed models, fully demonstrating the efficacy of NAS-Count.
\keywords{Crowd Counting, Neural Architecture Search, Multi-scale}
\end{abstract}
\section{Introduction}

Crowd counting, aiming to predict the number of individuals in a scene, has wide applications in the real world and receives considerable attention \cite{Survey2018,Survey2015a,Survey2015b}. With advanced occlusion robustness and counting efficiency, counting-by-density \cite{FirstDmapCC,MCNN,ECCV2018SANet,Arxiv2018CompositionLoss} has become the method-of-choice over others related techniques \cite{2008DCC,2009DCC,RidgeRegressionCC,Shah2013RCC,RidgeRegressionCC}. These techniques estimate a pixel-level density map and count the crowd by summing over pixels in the given area.

Although effective, counting-by-density is still challenged with scale variations induced by perspective distortion. To address this problem, most methods \cite{MCNN,ECCV2018SANet,Arxiv2018LeveragingUnlabeld} employ deep Convolutional Neural Network (CNN) for exploiting multi-scale features to perform density estimation in multi-scaled scenes. In particular, different-sized filters are arranged in parallel in multiple columns to capture multi-scale features for accurate counting in \cite{MCNN,HydraCNN,SwitchingCNN}, while in \cite{ECCV2018SANet,Arxiv2018CompositionLoss,tednet}, different filters are grouped into blocks and then stacked sequentially in one column. At the heart of these solutions, multi-scale capability originates from the compositional nature of CNN \cite{GNNSurvey,randomNet,hu2020attentional}, where convolutions with various receptive fields are composed hierarchically by hand. However, these manual designs demand prohibitive expert-efforts.

We therefore develop a Neural Architecture Search (NAS) \cite{zophRL,realEA} based approach to automatically discover the multi-scale counting-by-density models. NAS is enabled by the compositional nature of CNN and guided by human expertise in designing task-specific search space and strategies. For vision tasks, NAS blooms with image-level classification \cite{zophcvpr18,pnas,realAAAI191,realAAAI192}, where novel architectures are found to progressively transform spatial details to semantically deep features. Counting-by-density is, however, a pixel-level task that requires spatial preserving architectures with refrained down-sampling strides. Accordingly, the successes of NAS in image classification are not immediately transferable to crowd counting. Although attempts have been made to deploy NAS in image segmentation for pixel-level classifications \cite{NASsegNIPS18,autoDeepLab,NasSegAuxCell}, they are still not able to address counting-by-density, which is a pixel-level regression task with scale variations across the inputs.

In our NAS-Count, we propose a counting-oriented NAS framework with specific search strategy, search space and supervision method to develop our Automatic Multi-Scale Network (AMSNet). First, to achieve a fast search speed, we adopt a differential one-shot search strategy \cite{DARTS,autoDeepLab,PCDARTS}, in which architecture parameters are jointly optimized with gradient-based optimizer. Second, we employ a counting-specific two-level search space \cite{fabricsNIPS16,autoDeepLab}. On the micro-level, multi-scale cells are automatically explored to extract and fuse multi-scale features sufficiently. Pooling operations are limited to preserve spatial information and dilated convolutions are utilized instead for receptive field enlargement. On the macro-level, multi-path encoder-decoder architectures are searched to fuse multi-scale features from different cells and produce a high-quality density map.Fully-convolutional encoder-decoder is the architecture-of-choice for pixel-level tasks \cite{SegUnet,SDN2,ECCV2018SANet}, and the multi-path variant can better aggregate features encoded at different scales \cite{Refinenet,tednet,multipathCC19}. However, previous differential one-shot search strategies \cite{DARTS,PCDARTS,pdarts} mainly concentrate on the single-path network and neglects the effect of feature aggregation, which cannot efficiently fuse multi-scale features from different stages and is not suitable for crowd counting task.
\begin{figure*}
  \centering
  \subfigure[]{
    \label{fig:AMN} 
    \includegraphics[width=2.2in]{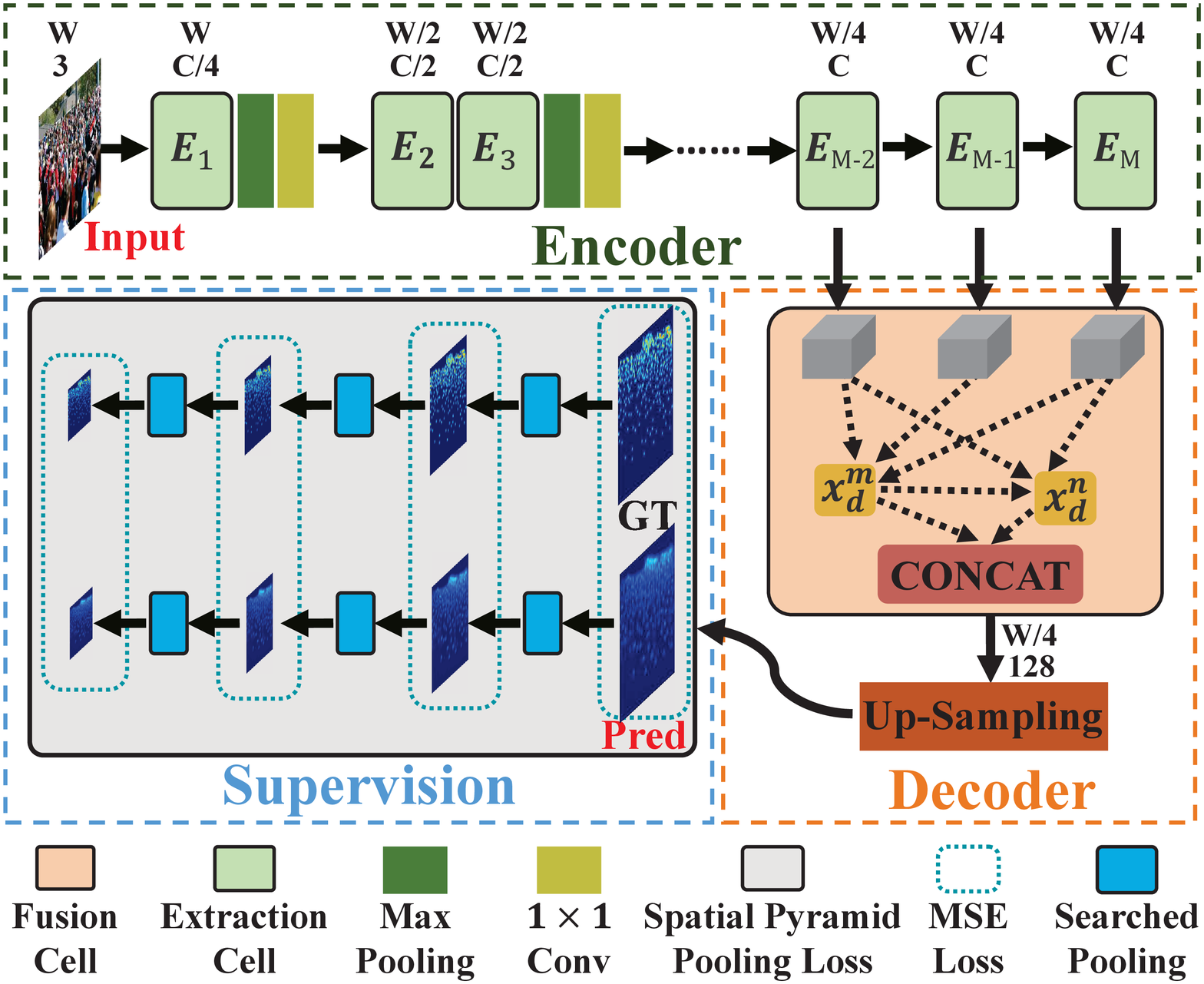}
  }
  \subfigure[]{
    \label{fig:searched_block} 
    \includegraphics[width=2.2in]{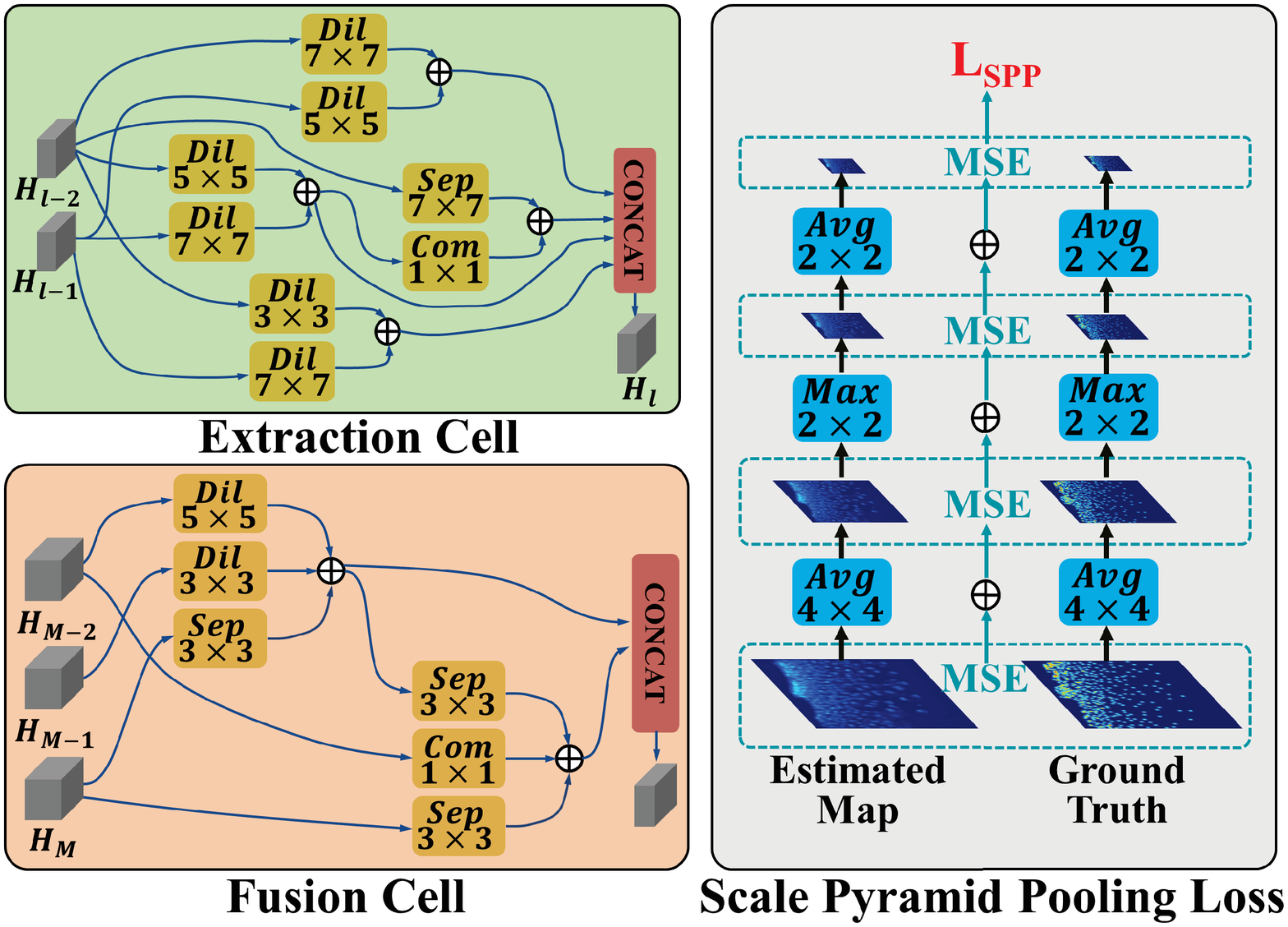}
  }
\vspace{-3mm}
  \caption{\textbf{(a):}An illustration of NAS-Count with the AMSNet architecture and SPPLoss supervision, all searched cells are outlined in black. Given $W \times W \times C$ ($C = 3$) inputs, the output dimension of each extraction and fusion cell are marked accordingly. \textbf{(b):}Detailed illustrations of the best searched cells. The circled additive sign denotes element-wise or scalar additions.}

\vspace{-6mm}
\end{figure*}
In our work, the multi-path exploration in macro-level search can solve this issue. Third, in order to address the pixel-level isolation problem \cite{ECCV2018SANet,CVPR2018CSRNet} of the traditional mean square error (MSE) loss, we propose to search an efficient Scale Pyramid Pooling Loss (SPPLoss) to optimize AMSNet. Leveraging the pyramidal pooling architecture to enforce supervision with multi-scale structural information has been prove effective in crowd counting task \cite{tednet,huang2018stacked,cheng2019learning}. However, its best internal components have not been explored well. Here, in our NAS-Count, we take a further step and automatically searched the best operation to extract multi-scale information in SPPLoss, which provides the more efficient supervision than manually designed one. By jointly searching AMSNet and SPPLoss, NAS-Count flexibly exploits multi-scale features and addresses the scale variation issue in counting-by-density. NAS-Count is illustrated in Figure \ref{fig:AMN}.

Main contributions of NAS-Count includes:
\vspace{-2mm}
\begin{itemize}
\setlength{\itemsep}{1pt}
\setlength{\parsep}{1pt}
\setlength{\parskip}{1pt}
\item To our best knowledge, NAS-Count is the first attempt at introducing NAS for crowd counting, where a multi-scale architecture is automatically developed to address the scale variation issue.
\item A counting-specific two-level search space is developed in NAS-Count, from which a multi-path encoder-decoder architecture (AMSNet) is discovered efficiently with a differentiable search strategy using stochastic gradient descent (SGD).
\item A Scale Pyramid Pooling Loss (SPPLoss) is searched automatically to improve MSE supervision, which helps produce the higher-quality density map via optimizing structural information on multiple scales.
\item By jointly searching AMSNet and SSPLoss, NAS-Count reports the best overall counting and density estimation performances on four challenging benchmarks, considerably surpassing other state-of-the-arts， which all require demanding expert-involvement.
\end{itemize}
\vspace{-2mm}

\section{Related work}
\label{sec:related}

\subsection{Crowd Counting Literature}
\label{sec:crowd}
Existing counting methods can be categorized into counting-by-detection \cite{2012HumanDetection,2009DCC,2008DCC,DCC2}, counting-by-regression \cite{RCC1,RCC2,Shah2013RCC,CNNpatch,MoCNN}, and counting-by-density strategies. For comprehensive surveys in crowd counting, please refer to \cite{Survey2015a,Survey2015b,Survey2018,beyondcounting}. The first strategy is vulnerable to occlusions due to the requirement of explicit detection. Counting-by-regression successfully avoids such requirement by directly regressing to a scalar count, but forfeits the ability to perceive the localization of crowds. The counting-by-density strategy, initially introduced in \cite{FirstDmapCC}, counts the crowd by first estimating a density map using hand-crafted \cite{FirstDmapCC,DensityCCNoDL2} or deep CNN \cite{MCNN,CVPR2018CSRNet,CCcvpr191,CCcvpr192} features, then summing over all pixel values in the map. Being a pixel-level regression task, CNN architectures deployed in counting-by-density methods tend to follow the encoder-decoder formulation. In order to handle scale variations with multi-scale features, single-column \cite{ECCV2018SANet,Arxiv2018CompositionLoss,ourcc} and multi-column \cite{MCNN,Crowdnet,HydraCNN,CNNboost} encoders have been used where different-sized convolution kernels are sequentially or parallelly arranged to extract features. For the decoder, hour-glass architecture with a single decoding path has been adopted \cite{ECCV2018SANet,Arxiv2018CompositionLoss,CCcvpr193}, while a novel multi-path variant is gaining increasing attention for superior multi-scale feature aggregation \cite{tednet,multipathCC19,Refinenet2,NasSegAuxCell}.

\vspace{-4mm}
\subsection{NAS Fundamentals}
\label{sec:neural}
\vspace{-2mm}
Although CNN have made great progress and achieved convincing performance in many computer vision tasks \cite{he2016deep,deeplab,hu2019two,jiang2019model}, its inherent structure often relies on the manual design, which demands enormous manpower and time. NAS, aiming to automatically explore the best structure of the network, has received considerable attention in recent years. The general efforts of developing new NAS solutions focus on designing new search spaces and search strategies. For search space, existing methods can be categorized into searching the network (macro) space \cite{realEA,zophRL}, the cell (micro) space \cite{zophcvpr18,pnas,realAAAI191,DARTS,PhamArXiv18}, or exploring such a two-level space \cite{fabricsNIPS16,autoDeepLab} jointly. The cell-based space search is the most popular where the ensemble of cells in networks is hand-engineered to reduce the exponential search space for fast computation. For search strategy, it is essentially an optimizer to find the best architecture that maximizes a targeted task-objective. Random search \cite{Random2012,Vizier}, reinforcement learning \cite{zophRL,zophcvpr18,MNasNet,RLAAAI18,RLArXiv16}, neuro-evolutionary algorithms \cite{EA2002,realEA,EA2019,EA2017ArXiv,realAAAI191,GeneticCNN}, and gradient-based methods \cite{DARTS,SNAS,ProxylessNAS} have been used to solve the optimization problem, but the first three suffer from prohibitive computation costs. Although many attempts have been made such as parameter sharing \cite{PhamArXiv18,paramshare17,RLAAAI18,paramshareICLM18}, hierarchical search \cite{pnas,autoDeepLab}, deploying proxy tasks with cheaper search space \cite{zophcvpr18} and training procedures \cite{predict1} to accelerate them, yet they are still far less efficient than gradient-based methods. Gradient-based NAS, represented by DARTS \cite{DARTS}, follows the one-shot strategy \cite{OneShotsmash} wherein a hyper-graph is established using differentiable architectural parameters. Based on the hyper-graph, an optimal sub-graph is explored within by solving a bi-level optimization with gradient-descent optimizers.
\vspace{-4mm}
\subsection{NAS Applications}
\label{sec:nasApp}
\vspace{-1mm}
NAS has shown great promise with discovered recurrent or convolutional neural networks in both sequential language modeling \cite{languageNAS} and multi-level vision tasks. In computer vision, NAS has excelled at image-level classification tasks \cite{zophcvpr18,realAAAI192,pnas,realAAAI191}, which is a customary starting-point for developing new classifiers outputting spatially coarsened labels. NAS was later extended to both bounding-box and pixel-level tasks, represented by object detection \cite{nasFPN,detNAS,autoFPN} and segmentation \cite{NASsegNIPS18,autoDeepLab,NasSegAuxCell}, where the search spaces are modified to better preserve the spatial information in the feature map. In \cite{NASsegNIPS18} a pixel-level oriented search space and a random search NAS were introduced to the pixel-level segmentation task. In \cite{NasSegAuxCell}, a similar search space was adopted, but the authors employed a reinforcement learning based search method. Nonetheless, both two methods suffer from formidable computations and are orders of magnitude slower than NAS-Count. In \cite{autoDeepLab}, the authors searched a two-level search space with more efficient gradient-based method, yet it dedicates in solving the pixel-level classification in semantic segmentation, which still differs from the per-pixel regression in counting-by-density.

\vspace{-2mm}
\section{NAS-Count Methodology}
\label{sec:AMN}
\vspace{-2mm}
NAS-Count efficiently searches a multi-scale encoder-decoder network, the Automatic Multi-Scale Network (AMSNet) as shown in Figure~\ref{fig:AMN}, in a counting-specific search space. It is then optimized with a jointly searched Scale Pyramid Pooling Loss (SPPLoss). The encoder and decoder in AMSNet consist of searched multi-scale feature extraction cells and multi-scale feature fusion cells, respectively, and SPPLoss deploys a two-stream pyramidal pooling architecture where the pooling cells are searched as well. By searching AMSNet and SPPLoss together, the operations searched in these two architectures can collaborate with each other to obtain the ideal multi-scale capability for addressing the scale-variation problem in crowd counting. NAS-Count details are discussed in the following subsections.
\vspace{-2mm}
\subsection{Automatic Multi-Scale Network}
\label{sec:search}
AMSNet is searched with the differential one-shot strategy in a two-level search space. To improve the search efficiency, NAS-Count adopts a continuous relaxation and partial channel connection as described in \cite{PCDARTS}. Differently, to alter the single-path formulation in \cite{PCDARTS}, we utilize the macro-level search to explore a multi-path encoder-decoder formulation for sufficient multi-scale feature extraction and fusion.

\vspace{-5mm}
\subsubsection{AMSNet Encoder}
The encoder of AMSNet is composed of a set of multi-scale feature extraction cells. For the $l$-th cell in the encoder, it takes the outputs of previous two cells, feature maps ${x_{l - 2}}$ and ${x_{l - 1}}$, as inputs and produces an output feature map ${x_l}$. We define each {\it cell} as a directed acyclic graph containing $N_{e}$ {\it nodes}, {\it i.e.} $x_e^i$ with $1 \leqslant i \leqslant {N_e}$, each represents a propagated feature map. We set ${N_e}$=7 containing two input nodes, four intermediate nodes, and one output node. Each directed {\it edge} in a cell indicates a convolutional operation $o_{e}(*)$ performed between a pair of nodes, and $o_{e}(*)$ is searched from the search space ${O_e}$ with 9 operations:
\vspace{-2mm}
\begin{itemize}
\setlength{\itemsep}{0pt}
\setlength{\parsep}{0pt}
\setlength{\parskip}{0pt}
\item $1 \times 1$ common convolution;
\item $3 \times 3$, $5 \times 5$, $7 \times 7$ dilated convolution with rate 2;
\item $3 \times 3$, $5 \times 5$, $7 \times 7$ depth-wise separable convolution;
\item skip-connection;
\item no-connection (zero);
\end{itemize}
\vspace{-2mm}

For preserving spatial fidelity in the extracted features, extraction cell involves no down-sampling operations. To compensate for the receptive field enlargement, we utilize dilated convolutions to substitute for the normal ones. Besides, we adopt depth-wise separable convolutions to keep the searched architecture parameter-friendly. Skip connections instantiate the residual learning scheme, which helps to improve multi-scale capacity as well as enhance gradient flows during back-propagation.

Within each cell, a specific intermediate node $x_e^m$ is connected to all previous nodes $x_e^1$, $x_e^2$ $ \cdots $, $x_e^{m - 1}$. Edges $o_{e}^{n,m}(*)$ are established between every pair of connected-nodes $n$ and $m$, forming a densely-connected hyper-graph. On a given edge $o_{e}^{n,m}(*)$ in the graph, following the continuously-relaxed differentiable search as discussed in \cite{DARTS}, its associated operation is defined as a summation of all possible operations weighted by the architectural parameter $\alpha_{e}$:
\begin{equation}
\begin{aligned}
o_e^{n,m}\left( {x_e^n;S} \right) = \sum\limits_i \sigma  (\alpha _e^{n,m,i}) \cdot o_e^i\left( {S \cdot x_e^n} \right) + \left( {1 - S} \right) \cdot x_e^n,\\
\end{aligned}
\label{equ:expropa}
\end{equation}
in the above equation, $\sigma(*)$ is a softmax function and $i = 9$ indicates the volume of the micro-level search space. Vector $S$ is applied to perform a channel-wise sampling on $x_e^n$, where 1/$K$ channels are randomly selected to improve the search efficiency. $K$ is set to 4 as proposed in \cite{PCDARTS}. $\alpha_{e}^{n,m}$ is a learnable parameter denoting the importance of each operation on an edge $o_{e}^{n,m}(*)$.

In addition, each edge is also associated with another architecture parameter $\beta _e^{n,m}$ which indicates its importance. Accordingly, an intermediate node $x_e^m$ is computed as a weighted sum of all edges connected to it:
\begin{equation}
x_e^m = \sum\limits_{n < m} \sigma  (\beta _e^{n,m}) \cdot o_e^{n,m}\left( {x_e^n;S} \right)
\label{equ:exedge}
\end{equation}
here, $n$ includes all previous nodes in the cell. The output of the cell is a concatenation of all its intermediate nodes. The cell architecture is determined by two architectural parameters $\alpha_{e}$ and $\beta _e$, which are jointly optimized with the weights of convolutions through a bi-level optimization. For details please refer to \cite{DARTS}. To recover a deterministic architecture from continuous relaxation, the most important edges as well as their associated operations are determined by computing $argmax$ on the product of $\sigma \left( {{\beta _e}} \right)$ and corresponding $\sigma \left( {{\alpha _e}} \right)$.

In the encoder, we apply a $1 \times 1$ convolution to preliminary encode the input image into a $\frac{C}{4}$ channel feature map. Afterwards, two $1 \times 1$ convolutions are implemented after the first and third extraction cells, each doubling the channel dimension of the features. Our searched extraction cell is normal cell that keeps the feature channel dimension unchanged. Spatially, we only reduce the feature resolution twice through two max pooling layers, aiming to preserve the spatial fidelity in the features, while double the channels before the two down-sampling operations. Additionally, within each extraction cell, an extra $1 \times 1$ convolution is attached to each input node, adjusting their feature channels to be one-fourth of the cell final output dimension.

\vspace{-2mm}
\subsubsection{AMSNet Decoder}
The decoder of AMSNet deploys a multi-scale feature fusion cell followed by an up-sampling module. We construct the hyper-graph of the fusion cell as inputting multiple features while outputting just one, therefore conforming to the aggregative nature of a decoder. The search in this hyper-graph is similar to that of the extraction cell. A fusion cell takes three encoder output feature maps as input, consisting of ${N_f}=6$ nodes that include three input nodes, two intermediate nodes and one output node. After the relaxation as formulated in Eqa.\ref{equ:expropa} and \ref{equ:exedge}, the architecture of a fusion cell is determined by its associated architecture parameters $\alpha_{d}$ and $\beta _d$. By optimizing $\beta _d$ on three edges connecting the decoder with three extraction cells in the encoder, NAS-Count fully explores the macro-level architecture of AMSNet, such that different single- or multi-path encoder-decoder formulations are automatically searched to discover the best feature aggregation for producing high-quality density maps. Through this macro-level search, we extend PC-DARTS from the single-path search strategy to a newly multi-path search strategy, which is more suitable for discovering a multi-scale network for crowd counting task.

As shown in Figure~\ref{fig:AMN}, $M$ denotes the number of extraction cells in the encoder and $C$ is the number of channels in the output of the last cell. To improve efficiency, we first employ a smaller proxy network, with $M$=6 and $C$=256, to search the cell architecture. Upon deployment, we enlarge the network to $M$=8 and $C$=512 for better performance. Through the multi-scale aggregation in the decoder, we obtain a feature map with 128 channels, which is then processed by an up-sampling module containing two $3 \times 3$ convolutions interleave with the nearest neighbor interpolation layers. The output of the up-sampling module is a single-channel density map with restored spatial resolution, which is then utilized in computing the SPPLoss.

\vspace{-3mm}
\subsection{Scale Pyramid Pooling Loss}
The default loss function to optimize counting-by-density models is the per-pixel mean square error (MSE) loss. By supervising this $L_{2}$ difference between the estimated density map and corresponding ground-truth, one assumes strong pixel-level isolation, such that it fails to reflect structural differences in multi-scale regions \cite{ECCV2018SANet,CVPR2018CSRNet}. As motivated by the Atrous Spatial Pyramid Pooling (ASPP) module designed in \cite{deeplab}, previous work \cite{tednet} attempts to solve this problem by proposing a new supervision architecture where non-parametric pooling layers are stacked into a two-stream pyramid. We call this supervision as Scale Pyramid Pooling Loss (SPPLoss). As shown in Figure~\ref{fig:searched_block}, after feeding the estimated map $E$ and the ground-truth $G$ into each stream, they are progressively coarsened and MSE losses are calculated on each level between the pooled maps. This is equivalent to computing the structural difference with increasing region-level receptive fields, and can therefore better supervise the pixel-level estimation model on different scales.

Instead of setting the pooling layers manually as in \cite{tednet}, NAS-Count searches the most effective SPPLoss architecture jointly with AMSNet. In this way, the multi-scale capability composed in both architecture can better collaborate to resolve the scale variation problem in counting-by-density. Specifically, each stream in SPPLoss deploys ${N_l}$=4 cascaded nodes. Among them, one input node is the predicted density map (or the given ground-truth). The other three nodes are produced through three cascaded searched pooling layers. The search space for operation $O_{l}$ performed on each edge contains six different pooling layers including:

\vspace{-2mm}
\begin{itemize}
\setlength{\itemsep}{0pt}
\setlength{\parsep}{0pt}
\setlength{\parskip}{0pt}
\item $2 \times 2$, $4 \times 4$, $6 \times 6$ max pooling layer with stride 2;
\item $2 \times 2$, $4 \times 4$, $6 \times 6$ average pooling layer with stride 2;
\end{itemize}
\vspace{-2mm}

The search for SPPLoss adopts the similar differentiable strategy as detailed in Section \ref{sec:search}. Notably, as SPPLoss is inherently a pyramid, its macro-level search space takes a cascaded form instead of a densely-connected hyper-graph. Accordingly, we only need to optimize the operation-wise architecture parameter $\alpha_{s}$ as follows:

\begin{equation}
o_s^{n,m}\left( {x_s^n} \right) = \sum\limits_i {\sigma \left( {\alpha _s^{n,m,i}} \right)}  \cdot o_s^i\left( {x_s^n} \right)
\label{equ:spppropa}
\end{equation}
$i$ indicates 6 different pooling operations, and ${x_s^n}$ represents an estimated map $E$ or ground-truth $G$ in specific level. Since both of them only have one channel, we thus do not apply partial channel connections ({\it i.e.} set $K$ equals to 1). The same cascaded architecture is shared in both streams of SPPLoss. Using the best searched architecture as depicted in Figure \ref{fig:searched_block}, SPPLoss is computed as:
\begin{equation}
L_{SPP} = \sum\limits_n \frac{1}{{{N^l}}}\left\| {\phi^l(E) - \phi^l(G)} \right\|_2^2
\label{equ:rclper}
\end{equation}
$N^l$ denotes the number of pixels in the map, $\phi^l(*)$ indicates the searched pooling operation, superscript $l$ is the layer index ranging from 0 to 3. $l=0$ is the special case where MSE loss is computed directly between $E$ and $G$.

\vspace{-4mm}
\section{Experiments}
\label{sec:experiment}
\vspace{-2mm}

\subsection{Implementation Details}
\label{sec:details}
The original annotations provided by the datasets are coordinates pinpointing the location of each individual in the crowd. To soften these hard regression labels for better convergence, we apply a normalized 2D Gaussian filter to convert coordinate map into density map, on which each individual is represented by a Gaussian response with radius equals to 15 pixels \cite{ourcc}.

\vspace{-3mm}
\subsubsection{Architecture Search}
The architecture of AMSNet and SPPLoss, {\it i.e.} their corresponding architecture parameters $\alpha_{{e,d,s}}$ and $\beta_{e,d}$, are jointly searched on the UCF-QNRF \cite{Arxiv2018CompositionLoss} training set. We choose to perform search on this dataset as it has the most challenging scenes with large crowd counts and density variations, and the search costs approximately 21 TITAN Xp GPU hours. Benefiting from the continuous relaxation, we optimize all architecture parameters and network weights $w$ jointly using gradient descent. Specifically, the first-order optimization proposed in \cite{DARTS} is adopted, upon which $w$ and $\alpha$, $\beta$ are optimized alternatively. For architecture parameters, we set the learning rate to be 6e-4 with weight decay of 1e-3. We follow the implementation as in \cite{PCDARTS,autoDeepLab}, where a warm-up training for network weights is first conducted for 40 epochs and stops the search early at 80 epochs. For training the network weights, we use a cosine learning rate that decays from 0.001 to 0.0004, and weight decay 1e-4. Data augmentation including random-scale sampling, random flip and random rotation are conducted to alleviate overfitting.

\begin{figure}[t]
\begin{center}
\includegraphics[width=3.0in]{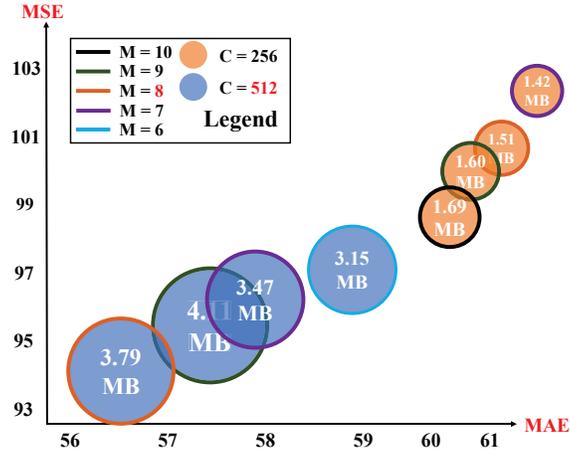}
\vspace{-5mm}
\end{center}
   \caption{Illustrated hyper-parameter analysis. $M$ is the number of extraction cells, $C$ denotes the channels of feature map generated by the last extraction cell. Bottom left corner indicates superior counting result and the number in the circle indicates the parameter size of each model. The best hyper-parameters are colored with red in the legend.}
\label{fig:ablation}
\vspace{-4mm}
\end{figure}

\vspace{-3mm}
\subsubsection{Architecture Training}
After the architectures of AMSNet and SPPLoss are determined by searching on the UCF-QNRF dataset, we re-train the network weights $w$ from scratch on each dataset respectively. We re-initialize the weights with Xavier initialization, and employ Adam with initial learning rate set to 1e-3. This learning rate is decayed by 0.8 every 15K iterations.

\vspace{-3mm}
\subsubsection{Architecture Evaluation}
Upon deployment, we directly feed the whole image into AMSNet, aiming to obtain high-quality density maps free from boundary artifacts. In counting-by-density, the crowd count on an estimated density map equals to the summation of all pixels. To evaluate the counting performance, we follow the previous work and employ the widely used mean average error (MAE) and the mean squared error (MSE) metrics. Additionally, we also utilize the PSNR (Peak Signal-to-Noise Ratio) and SSIM (Structural Similarity in Image) metrics to evaluate density map quality \cite{PyramidCNNsICCV2017}.

\begin{table*}[t]
\setlength{\tabcolsep}{4mm}
\caption{Estimation errors on the ShanghaiTech. The best performance is colored red and the second best is colored blue.}
\vspace{-3mm}
\begin{center}
\small
\begin{tabular}{|l|cc|cc|}
\hline
\multirow{2}{*}{Method} &  \multicolumn{2}{|c|}{ShanghaiTech Part\_A} & \multicolumn{2}{|c|}{ShanghaiTech Part\_B} \\
\cline{2-5}
& MAE$\downarrow$ & MSE$\downarrow$  & MAE$\downarrow$ & MSE$\downarrow$  \\
\hline
\hline
MCNN \cite{MCNN} & 110.2& 173.2& 26.4& 41.3\\
CSRNet \cite{CVPR2018CSRNet}& 68.2& 115.0& 10.6& 16.0\\
SANet \cite{ECCV2018SANet}& 67.0& 104.5& 8.4& 13.6\\
CFF \cite{shi2019counting}& 65.2& 109.4& 7.2& 12.2\\
TEDNet \cite{tednet} & 64.2& 109.1& 8.2& 12.8\\
SPN+L2SM \cite{xu2019learn}& 64.2& 98.4& 7.2& 11.1\\
ANF \cite{zhang2019attentional} & 63.9& 99.4& 8.3& 13.2\\
PACNN+ \cite{shi2019revisiting} & 62.4& 102.0& 7.6& 11.8\\
CAN \cite{CCcvpr192} & 62.3& 100.0& 7.8& 12.2\\
SPANet \cite{cheng2019learning} & 59.4& \textcolor{blue}{92.5}& \textcolor{red}{6.5}& \textcolor{red}{9.9}\\
PGCNet \cite{yan2019perspective} & \textcolor{blue}{57.0}& \textcolor{red}{86.0}& 8.8& 13.7\\
\hline
AMSNet & \textcolor{red}{56.7}& 93.4& \textcolor{blue}{6.7}& \textcolor{blue}{10.2}\\
\hline
\end{tabular}
\end{center}
\label{table:comparison1}
\vspace{-8mm}
\end{table*}

\begin{table*}[t]
\setlength{\tabcolsep}{4mm}
\caption{Estimation errors on the UCF\_CC\_50 and the UCF-QNRF datasets. The best performance is colored red and the second best is colored blue.}
\vspace{-3mm}
\begin{center}
\small
\begin{tabular}{|l|cc|cc|}
\hline
\multirow{2}{*}{Method} & \multicolumn{2}{|c|}{UCF\_CC\_50} & \multicolumn{2}{|c|}{UCF-QNRF}\\
\cline{2-5}
& MAE$\downarrow$ & MSE$\downarrow$  & MAE$\downarrow$ & MSE$\downarrow$     \\
\hline
\hline
Zhang {\it et al}. \cite{CNNpatch}& 467.0& 498.5& $\_$ & $\_$\\
MCNN \cite{MCNN} & 377.6& 509.1& 277& 426\\
CP-CNN \cite{PyramidCNNsICCV2017}& 295.8& 320.9& $\_$ & $\_$\\
CSRNet \cite{CVPR2018CSRNet}& 266.1& 397.5& $\_$ & $\_$\\
SANet \cite{ECCV2018SANet}& 258.4& 334.9& $\_$ & $\_$\\
TEDNet \cite{tednet} & 249.4& 354.5& 113& 188\\
ANF \cite{zhang2019attentional} & 250.2& 340.0& 110 & 174\\
PACNN+ \cite{shi2019revisiting} & 241.7& 320.7& $\_$ & $\_$\\
CAN \cite{CCcvpr192} & 212.2& \textcolor{red}{243.7}& 107& 183\\
CFF \cite{shi2019counting}& $\_$ & $\_$& \textcolor{red}{93.8}& \textcolor{red}{146.5}\\
SPN+L2SM \cite{xu2019learn} & \textcolor{red}{188.4}& 315.3& 104.7& 173.6\\
\hline
AMSNet & \textcolor{blue}{208.4}& \textcolor{blue}{297.3}& \textcolor{blue}{101.8} & \textcolor{blue}{163.2}\\
\hline
\end{tabular}
\end{center}
\label{table:comparison2}
\vspace{-8mm}
\end{table*}

\begin{table}[t]
\setlength{\tabcolsep}{2mm}
\caption{The MAE comparison on WorldExpo'10. The best performance is colored red and second best is colored blue.}
\vspace{-3mm}
\begin{center}
\begin{tabular}{|l|c|c|c|c|c|c|}
\hline
Method & S1 & S2 & S3 & S4 & S5 & Ave.\\
\hline
\hline
SANet \cite{ECCV2018SANet} & 2.6 & 13.2 & 9.0 & 13.3 & 3.0 & 8.2\\
CAN \cite{CCcvpr192} & 2.9 & 12.0& 10.0& \textcolor{red}{7.9}& 4.3& 7.4\\
DSSIN \cite{liu2019crowd} & \textcolor{red}{1.6} & 9.5& 9.5& \textcolor{blue}{10.4}& \textcolor{red}{2.5}& \textcolor{red}{6.7}\\
ECAN \cite{CCcvpr192} & 2.4 & \textcolor{blue}{9.4}& \textcolor{blue}{8.8}& 11.2& 4.0& 7.2\\
TEDNet \cite{tednet} & 2.3 & 10.1 & 11.3 & 13.8 & \textcolor{blue}{2.6} & 8.0\\
AT-CSRNet \cite{CCcvpr193} & \textcolor{blue}{1.8} & 13.7 & 9.2 & 10.4 & 3.7 & 7.8\\
ADMG \cite{wan2019adaptive} & 4.0 & 18.1& \textcolor{red}{7.2}& 12.3& 5.7& 9.5\\
\hline
AMSNet & \textcolor{red}{1.6} & \textcolor{red}{8.8}& 10.8& \textcolor{blue}{10.4}& \textcolor{red}{2.5}& \textcolor{blue}{6.8}\\
\hline
\end{tabular}
\end{center}
\label{table:wecomparison}
\vspace{-4mm}
\end{table}

\subsection{Search Result Analysis}
\label{sec:searchre}
The best searched multi-scale feature extraction and fusion cells, as well as the SPPLoss architecture are illustrated in Figure \ref{fig:searched_block}. As shown, extraction cell maintains the spatial and channel dimensions unchanged ($1 \times 1$ convolutions are employed to manipulate the channel dimensions in the cells). The extraction cell primarily exploits dilated convolutions over normal ones, conforming to the fact that in the absence of heavy down-samplings, pixel-level models rely on dilations to enlarge receptive fields. Furthermore, different kernel sizes are employed in the extraction cell, showing its multi-scale capability in addressing scale variations. By taking in three encoded features and generating one output feature, the fusion cell constitutes a multi-path decoding hierarchy, wherein primarily non-dilated convolutions with smaller kernels are selected to aggregate features more precisely and parameter-friendly.

\subsection{Ablation Study on Searched Architectures}
\label{sec:effsearched}
For ablation purposes, we employ the architecture proposed in \cite{ECCV2018SANet} as the baseline encoder (composed of four inception-like blocks). Additionally, to better elaborate the effectiveness of the search process, we also employ the backbone searched on ImageNet in \cite{PCDARTS} to compose the classification encoder (For the consideration of computation cost and fair comparison, we totally set 8 cells in encoder, which is the same in our AMSNet). The baseline decoder cascades two $3 \times 3$ convolutions interleaved with nearest-neighbor interpolation layers. The normal MSE loss is utilized as baseline supervision. By comparing different modules with its baseline, the ablation study results on the ShanghaiTech Part\_A dataset are reported in Table \ref{table:component}. This table is partitioned into three groups, and each row indicates a specific configuration. The MAE and PSNR metrics are used to show the counting accuracy and density map quality.

\begin{table}
\caption{Ablation study results. Best performance is bolded, and arrows indicate the favorable directions of the metric values.}
\vspace{-3mm}
\begin{center}
\small
\begin{tabular}{|c|c|c|c|c|}
\hline
\multicolumn{3}{|c|}{Configurations} & MAE$\downarrow$ & PSNR$\uparrow$ \\
\hline
\multirow{6}{*}{\makecell[tl]{Encoder \\ Architecture}}
&\multirow{2}{*}{1} & Baseline Encoder & \multirow{2}{*}{69.1} & \multirow{2}{*}{23.54} \\
& &Baseline Decoder & & \\
\cline{2-5}
&\multirow{2}{*}{2}&Classification Encoder & \multirow{2}{*}{{67.8}} & \multirow{2}{*}{{23.67}} \\
& &Baseline Decoder & & \\
\cline{2-5}
&\multirow{2}{*}{3}&AMSNet Encoder & \multirow{2}{*}{\textbf{60.3}} & \multirow{2}{*}{\textbf{25.82}} \\
& &Baseline Decoder & & \\ \hline
\multirow{4}{*}{\makecell[tl]{Decoder \\  Architecture}}
&\multirow{2}{*}{1}& Baseline Encoder & \multirow{2}{*}{69.1} & \multirow{2}{*}{23.54} \\
& & Baseline Decoder & & \\ 
\cline{2-5}
&\multirow{2}{*}{4}& Baseline Encoder & \multirow{2}{*}{\textbf{62.4}} & \multirow{2}{*}{\textbf{24.75}} \\ 
& & AMSNet Decoder & & \\ \hline
\multirow{3}{*}{Supervision}
&5& AMSNet + MSE & 58.5 & 26.17\\
\cline{2-5}
&6& AMSNet + SAL & 57.6 & 26.62 \\
\cline{2-5}
&7& AMSNet + SPPLoss & \textbf{56.7} & \textbf{27.03} \\ \hline
\end{tabular}
\end{center}
\label{table:component}
\vspace{-8mm}
\end{table}

\begin{table}[t]
\setlength{\tabcolsep}{2mm}
\caption{Model size and performance comparison among state-of-the-art counting methods on the ShanghaiTech Part\_A.}
\vspace{-3mm}
\begin{center}
\small
\begin{tabular}{|l|c|c|c|c|}
\hline
Method  & MAE$\downarrow$ & PSNR$\uparrow$ &SSIM$\uparrow$ & Size\\
\hline
\hline
MCNN \cite{MCNN}& 110.2 & 21.4 & 0.52 & \textbf{0.13MB}\\
Switch-CNN \cite{SwitchingCNN} & 90.4 & $\_$ & $\_$ & 15.11MB\\
CP-CNN \cite{PyramidCNNsICCV2017} & 73.6 & 21.72 & 0.72 & 68.4MB\\
CSRNet \cite{CVPR2018CSRNet} & 68.2 & 23.79 & 0.76 & 16.26MB\\
SANet \cite{ECCV2018SANet}& 67.0 & $\_$ & $\_$ & 0.91MB\\
TEDNet \cite{tednet}& 64.2  & 25.88 & 0.83 & 1.63MB\\
ANF \cite{zhang2019attentional}& 63.9 & 24.1 & 0.78 & 7.9MB\\\hline
AMSNet & \textbf{56.7} & \textbf{27.03} & \textbf{0.89} & 3.79MB\\
AMSNet\_light & 61.3 & 26.18 &0.85 & 1.51MB\\
\hline
\end{tabular}
\end{center}
\label{table:parameters}
\vspace{-8mm}
\end{table}

Architectures in the first two groups (five rows) are optimized with the normal MSE loss. As shown, compared to the baseline, the searched AMSNet encoder improves counting accuracy and density map quality by 12.7\% and 9.7\%, while the searched decoder brings 9.7\% and 5.1\% improvements respectively. Meanwhile, compared to the classification encoder, AMSNet encoder also improves the performance by 11.1\% and 9.1\% in MAE and PSNR, which indicates we obtain a more powerful backbone for multi-scale feature extraction through the search process. In the third group, AMSNet is supervised by different loss functions to demonstrate their efficacy. The Spatial Abstraction Loss (SAL) proposed in \cite{tednet} adopts a hand-designed pyramidal architecture, which surpasses the normal MSE supervision on both counting and density estimation performance. These improvements are further enhanced by deploying SPPLoss, showing that the searched pyramid benefits counting and density estimation by supervising multi-scale structural information.

Furthermore, we also compare AMSNet decoder with some existing multi-path decoder to show the ability of our macro-level search in discovering an efficient feature aggregation configuration. These experiments are elaborated in detail in the supplementary material.

\subsection{Hyper-parameter Study}
\label{sec:comhyper}
The size and performance of AMSNet are largely dependent on two hyper-parameter $M$ and $C$, each denoting the number of extraction cell and its output channel dimension. As illustrated in Figure \ref{fig:ablation}, $M=8$ and $C=512$ render the best counting performance, but populate AMSNet with 3.79MB parameters. When decreasing $C$ to 256, the size of AMSNet also shrinks dramatically, but at the expense of decreased accuracy. Nevertheless, $M=8$ still produces the best MAE in this case. As a result, we configure our AMSNet with $M=8, C=512$, and also establish an AMSNet\_light with $M=8, C=256$ in the experiment.

We compare the counting accuracy and density map quality of both AMSNet and AMSNet\_light with other state-of-the-art counting methods in Table \ref{table:parameters}. As shown, AMSNet reports the best MAE and PSNR overall, while being heavier than three other methods. AMSNet\_light, on the other hand, is the third most light model and achieves the best performance with the exception of AMSNet.

\vspace{-3mm}
\subsection{Performance and Comparison}
\label{sec:performance}
\vspace{-1mm}
We compare the counting-by-density performance of NAS-Count with other state-of-the-art methods on four challenging datasets, ShanghaiTech \cite{MCNN}, WorldExpo'10 \cite{CNNpatch}, UCF\_CC\_50 \cite{Shah2013RCC} and UCF-QNRF \cite{Arxiv2018CompositionLoss}. In particular, the counting accuracy comparison is reported in Tables \ref{table:comparison1}, \ref{table:comparison2} and \ref{table:wecomparison}, while the density map quality result is shown in Table \ref{table:parameters}.

\vspace{-3mm}
\subsubsection{Counting Accuracy}
\label{sec:countingaccu}
\vspace{-1mm}

The ShanghaiTech is composed of Part\_A and Part\_B with in total of 1198 images. It is one of the largest and most widely used datasets in crowd counting. As shown in Table \ref{table:comparison1}, AMSNet achieves the state-of-the-art performance in terms of both MAE and MSE. On Part\_A, we achieve the best MAE and the competitive MSE. On Part\_ B, we report the second best MAE and MSE, which are only a little inferior to \cite{cheng2019learning}.

The UCF\_CC\_50 dataset contains 50 images of varying resolutions and densities. In consideration of sample scarcity, we follow the standard protocol \cite{Shah2013RCC} and use 5-fold cross-validation to evaluate method performance. As shown in Table \ref{table:comparison2}, we achieve the second best MAE and MSE. It is worth mentioning that, although our MAE is a higher than SPN+L2SM \cite{xu2019learn}, our MSE is obviously better than it. Meanwhile, our MAE is also superior to CAN \cite{CCcvpr192}, which is the only current method achieves a lower MSE than our AMSNet. Therefore, AMSNet produces the best performance when we comprehensively consider both MAE and MSE together.

The UCF-QNRF dataset introduced by Idress {\it et al}. \cite{Arxiv2018CompositionLoss} has images with the highest crowd counts and largest density variation, ranging from 49 to 12865 people per image. These characteristics make UCF-QNRF extremely challenging for counting models. As shown in Table \ref{table:comparison2}, we achieve the second best performance in terms of both MAE and MSE on this dataset.

The WorldExpo'10 dataset \cite{CNNpatch} contains 3980 images covering 108 different scenes. As shown in Table \ref{table:wecomparison}, AMSNet achieves the second lowest average MAE over five scenes, and also performs the best on the three scenes individually.

It is worth noting that although we do not produce the best counting accuracy on every dataset. Our AMSNet is the only method that achieves the top-two performance on the four employed datasets simultaneously. In the other word, AMSNet performs best when we comprehensively consider the four datasets.

\begin{figure*}[t]
\begin{center}
\includegraphics[width=1\linewidth]{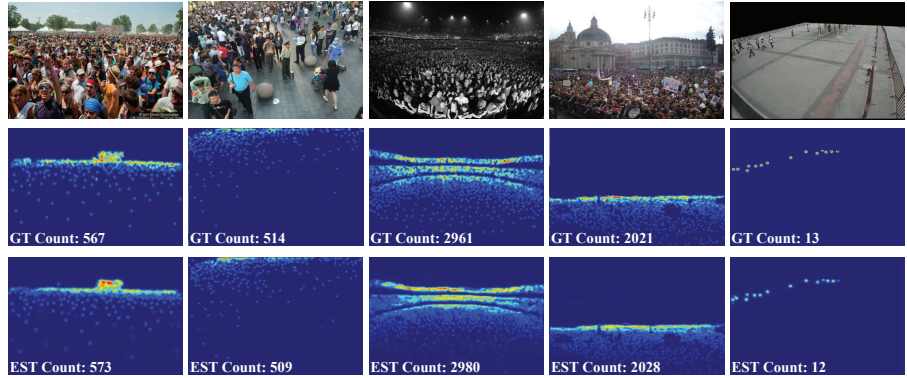}
\vspace{-8mm}
\end{center}
   \caption{An illustration of generated density maps on ShanghaiTech Part\_A, ShanghaiTech Part\_B, UCF\_50\_CC, UCF-QNRF and WorldExpo'10 respectively. The first row shows the input images, the second and third depict the ground truth and estimated density maps.}
\label{fig:selfvis}
\end{figure*}

\vspace{-3mm}
\subsubsection{Density Map Quality}
\label{sec:mapquality}
As shown in Table \ref{table:parameters}, we employ PSNR and SSIM indices to compare the quality of density maps estimated by different methods. AMSNet performs the best on both indices, outperforming the second best by 4.4\% and 7.2\% respectively. Notably, even by deploying AMSNet\_light which is the third lightest model, we still generate the most high-quality density map. We further showcase more density maps generated by AMSNet on all employed datasets in Figure \ref{fig:selfvis}.

\vspace{-2mm}
\section{Conclusion}
\label{sec:conclusion}
\vspace{-2mm}

NAS-Count is the first endeavor introducing neural architecture search into counting-by-density. In this paper, we extend PC-DARTS \cite{PCDARTS} to a counting-specific two-level search space, in which micro- and macro-level search are employed to explore a multi-path encoder-decoder network, AMSNet, as well as the SPPLoss. Specifically, AMSNet employs a novel composition of multi-scale feature extraction and fusion cells. SPPLoss automatically searches a scale pyramid architecture to extend normal MSE loss, which helps to supervise structural information in the density map at multiple scales. By jointly searching AMSNet and SPPLoss end-to-end, NAS-Count surpasses tedious hand-designing efforts by achieving a multi-scale model automatically with less than 1 GPU day, and demonstrates overall the best performance on four challenging datasets.

\section*{Acknowledgment}
This paper was supported by the National Natural Science Foundation of China (NSFC) under grant 91738301, and the National Key Scientific Instrument and Equipment Development Project under Grant 61827901.

\clearpage
%
%
\bibliographystyle{splncs04}
\bibliography{egbib}
\end{document}